\documentclass[runningheads,a4paper]{llncs}
\usepackage[pdftex]{graphicx}
\usepackage{amsmath,latexsym,amssymb}
\usepackage{hyperref}
\usepackage{booktabs}
\usepackage{caption}
\usepackage{tabularx}
\usepackage[usenames]{color}
\usepackage{indentfirst}
\usepackage{pbox}
\definecolor{purple}{rgb}{1,0,1}
\definecolor{lightgray}{rgb}{0.7,0.7,0.7}

\newcommand{\zm}[1]{\textcolor{black}{#1}}

\DeclareMathOperator*{\argmax}{arg\,max} 
\DeclareMathOperator*{\argmin}{arg\,min}

\newcommand{\eq}[1]{(#1)}
\newcommand{\fig}[1]{Fig.~#1}
\newcommand{\tab}[1]{Table~#1}

\newcommand{\cmmnt}[1]{}

\begin{document}
\title{\fontdimen2\font=4pt Star Shape Prior in Fully Convolutional Networks for Skin Lesion Segmentation}
\titlerunning{Star shape prior in FCNs}
\authorrunning{Mirikharaji, and Hamarneh.}
\author{Zahra Mirikharaji, Ghassan Hamarneh}

\institute{School of Computing Science, Simon Fraser University, Canada\\
\email{\{zmirikha, hamarneh\}@sfu.ca}\\}

\maketitle

\begin{abstract}

Semantic segmentation is an important preliminary step towards automatic medical image interpretation. Recently deep convolutional neural networks have become the first choice for the task of pixel-wise class prediction. While incorporating prior knowledge about the structure of target objects has proven effective in traditional energy-based segmentation approaches, there has not been a clear way for encoding prior knowledge into deep learning frameworks. In this work, we propose a new loss term that encodes the star shape prior into the loss function of an end-to-end trainable fully convolutional network (FCN) framework. We penalize non-star shape segments in FCN prediction maps to guarantee a global structure in segmentation results. Our experiments demonstrate the advantage of regularizing FCN parameters by the star shape prior and our results on the ISBI 2017 skin segmentation challenge data set achieve the first rank in the segmentation task among $21$ participating teams.

\end{abstract}

%

\section{Introduction}

Skin cancer is the most common type of cancer in the world. Early detection of skin cancer can increase the five year survival rate of patients from 18\% to 98\%~\cite{cancer2017}. While skin cancer can be detected by visual examination, distinguishing malignant from non-malignant lesions is a challenging task. In recent years, computer aided diagnosis has been widely leveraged in automated assessment of dermoscopy and clinical images to assist dermatologists evaluation. Semantic segmentation, the task of labeling each image pixel with the class label of its surrounding object, is generally the first step toward the automatic understanding of images. Remarkable variations in the appearance of healthy and unhealthy skin, including color, texture, lesion shape and size originating from image acquisition and inter- and intra-class variation, complicates the skin lesion segmentation problem. 

For decades, since the seminal work of Kass et al.~\cite{kass1988snakes}, energy functional minimization techniques were the most popular approaches to solve image segmentation problems~\cite{mcinerney1996deformable}. Imaging artifacts and variability in the appearance of image regions make the data fidelity term insufficient to achieve robust segmentation results. Therein, the segmentation that minimizes a weighted sum of unary (data) and regularization energy functional terms is sought. Incorporating prior knowledge about the structure of target object in the objective function to regularize plausible solutions with anatomically meaningful constraints have been widely leveraged to obtain more reliable delineations~\cite{cremers2007review,nosrati2016incorporating}. Active shape models (ASM) was one of the pioneering works to incorporate shape priors into deformable models~\cite{cootes2001active}. To effectuate the shape prior, ASM and many other shape-encoding segmentation methods required an estimate of the object pose (i.e., the orientation, scale, and location of the target object in the image)~\cite{freedman2005interactive,vu2008shape}. Some examples of priors which have been utilized in energy optimization based segmentation methods are shape models, topology preservation, moment constraints and geometrical and distance interaction between image regions.


Recently deep fully convolutional networks have achieved significant success in the task of semantic segmentation.  Hierarchical extraction of features followed by skip connections and up-sampling operations was first introduced by Long et al. in an end-to-end trainable framework~\cite{long2015fully}. Despite the success of FCNs, they have indicated clear limitations in the dense per-pixel prediction task. Consecutive spatial pooling and striding convolutions in FCNs reduce the initial image resolution and lead to loss of the image fine structures. Some techniques have been proposed to address these limitations of FCNs. Learning multiple deconvolutional layers and concatenating low-level fine features with high-level coarse features through skip connections are commonly used to retrieve low-level visual features~\cite{ronneberger2015u}. Dilated convolutional has also been introduced to aggregate multi-scale contextual information without losing image resolutions~\cite{yu2015multi}. Although pixel-wise prediction benefits from these resolution enlarging techniques, they are only capable to partially recover detailed spatial information. 

In the context of fully convolutional networks, leveraging prior information about the target object structure in the segmentation model has not been widely studied. By optimizing individual pixel level class predictions in the FCNs loss function, independent class labels are assigned to image pixels without considering high-level label dependencies. There have been some efforts towards structured prediction and leveraging meaningful priors into deep learning frameworks. Deeplab-CRF and CRF-RNN employ probabilistic graphical modeling either as a post processing step or by implementing recurrent layers in FCNs to enforce assigning similar labels to pixels with similar color and position and further improve the object boundaries~\cite{chen2016deeplab}. Recently BenTaieb et al. proposed a new loss function to encode the geometrical and topological priors of containment and detachment in an end-to-end FCN framework~\cite{bentaieb2016topology,zheng2015conditional}. To leverage the shape prior in segmentation models, Chen et al. learn a shape constraint by a deep Boltzmann machine and then employ the learned prior in a variational segmentation method~\cite{chen2013deep}. In addition, training convolutional auto-encoder networks to learn anatomical shape variations has demonstrated improvements in the robustness of FCN segmentation models~\cite{oktay2018anatomically,ravishankar2017learning}. 

To the best of our knowledge, none of the existing works incorporates a star shape prior as a regularization term in the loss function of FCNs trained in an end-to-end fashion. The star shape prior was first introduced in the context of image segmentation by Veksler, where it was encoded as a regularization term into the cost function formulation of a graph-based (discrete) image segmentation approach~\cite{veksler2008star}. Later, Chittajallu et al. incorporated three types of shape constraints including star shape prior into a Markov random field based segmentation model and applied their method to non-contrast cardiac computed tomography scans~\cite{chittajallu2010shape}. Yuan et al. extended the star shape prior to 3D objects and applied it to prostate magnetic resonance images~\cite{yuan2012efficient}. Nosrati et al. derived a star shape prior in a continuous variational formulation and applied it to segmenting overlapping cervical cells~\cite{nosrati2015segmentation}. Although the star shape prior clearly improved results for a variety of target objects, one limiting requirement of Veksler’s approach and its variants, however, is the assumption that the center of foreground objects is known (e.g. provided by user interaction).

We aim to harness the powerful proven capabilities of deep learning in automatically extracting learnt (i.e., not hand-crafted) pixel-driven image features (i.e., likelihood) and augment it with demonstrably useful shape priors without requiring the knowledge of the target object pose. We propose to encode the star shape prior into the training of fully convolutional networks to improve segmentation of skin lesions from their surrounding healthy skin.\cmmnt{ We train a deep segmentation model that generates prediction maps satisfying the star shape definition with respect to the skin lesion center.} Our idea is to formulate the star shape prior in the loss function of FCN frameworks to penalize non-star shape segments in prediction maps and preserve global structures in the output space. Integration of the star shape prior in the loss function makes it possible to train the whole FCN framework in an end-to-end manner. In contrast to Veksler's work and its variants, our approach to star shape prior in a deep learning setting not only eliminates the need for manually setting object centers, but also alleviates, at inference time, the computationally intensive optimization associated with the energy minimizing approaches. Our experimental results illustrate how imposing the shape prior constraint in deep networks refines skin lesion segmentation in comparison to using a single pixel level loss in FCNs. 

\section{Methodology}

Our goal is to leverage the star shape prior into the learning process of an FCN to generate plausible segmentation maps (e.g. skin lesions) from their surrounding background without requiring additional training, user interaction, pre- or post-processing. \\

\noindent\textbf{FCN's pixel-wise loss}~~In FCNs, given a set of $N$ training images and their corresponding ground truth segmentations, $\{(X(i), Y(i)); i = 1,2, \dots , N\}$, the deep network learns to take unseen image samples and generate a segmentation probability map, the same size as the input images that assigns a semantic label to each pixel. Learning the deep network parameters $\theta$, is performed by maximizing the a posteriori probability of giving the true label to each image pixel given the input image. Maximizing the a posteriori probability is usually replaced by minimizing its negative log-likelihood function as a cost function $L$:

\begin{equation}
\theta^*= \argmin_\theta L(X,Y;\theta).
\label{total_loss}
\end{equation}

\cmmnt{
\begin{equation}
\theta^*= \argmax_\theta \sum_{i=1}^{N} \sum_{p \in \Omega} P(y'_{ip}=y_{ip}|X(i);\theta).
\label{post_prob}
\end{equation}
}
For binary dense class prediction, a binary cross entropy loss $L_{ce}$ is generally deployed:
\begin{equation}
\begin{split}
L_{ce}(X, Y;\theta)=-\sum_{i=1}^{N} \sum_{p \in \Omega}~[y_{ip} \log P(y_{ip}=1|X(i);\theta)\\ + (1-y_{ip})\log (1-P(y_{ip}=1|X(i);\theta))]
\label{cross_entropy_loss}
\end{split}
\end{equation}

\noindent where \zm{$\Omega$ is the pixel space,} $y_{ip}$ is the ground truth label of pixel $p$ in image $i$ and $P(y_{ip}=1|X(i);\theta)$ is the FCN sigmoid function output indicating the predicted probability of the $p^{th}$ pixel of the $i^{th}$ image being a skin lesion. The pixel-wise binary logistic loss $L_{ce}$ penalizes the deviation of the predicted label for each pixel from its true label.\\

\begin{figure*}[!t]
\centering
\includegraphics[width=4.8in]{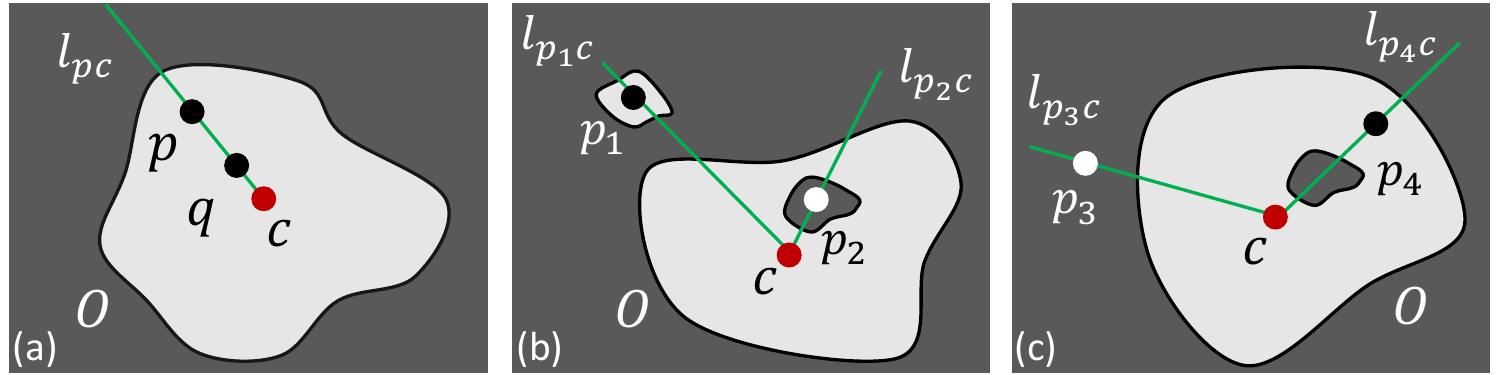}
\caption{(a) Star shape object $O$ w.r.t. the supplied object center $c$ (\emph{red dot}). (b) Examples of the star shape constraint violation. (c) Examples of cases where conditions (i) and (ii) in \eq{\ref{star_shape_loss}} are required.}
\vspace{-1em}
\label{starshape}
\end{figure*}

\noindent\textbf{Star shape regularized loss}~~Assuming $c$ is the center of object $O$, object $O$ is a star shape object if, for any point $p$ interior to the object, all the pixels $q$ lying on the straight line segment connecting $p$ to the object center $c$ are inside the object (\fig{\ref{starshape}}-(a)). This definition of star shape prior holds for a large group of object shapes including convex ones. To incorporate the star shape prior as a new regularization term, we augment the loss function in \eq{\ref{cross_entropy_loss}} with a new loss term to penalize line segments that violate the prior (e.g. \fig{\ref{starshape}}-(b)) in the prediction maps: 
\begin{eqnarray}
L(X,Y;\theta)=\alpha L_{ce} + \beta L_{sh}
\label{total_loss}
\end{eqnarray}
 where $\alpha$ and $\beta$ are hyper-parameters setting the contribution of each term in the optimization function, $L_{ce}$ is the binary cross entropy loss and $L_{sh}$ is our star shape prior:
\begin{equation}
\begin{split}
L_{sh}(X,Y;\theta)=\sum_{i=1}^{N}\sum_{p \in \Omega}\sum_{q \in l_{pc}} B_{pq}^{i}\times|y_{ip}-P (y_{ip}=1|X(i);\theta)|\\    \times |P(y_{ip}=1|X(i);\theta )-P(y_{iq}=1|X(i);\theta)|;
\label{star_shape_loss}
\end{split}
\end{equation}

\begin{equation}
B_{pq}^{i}= 
\begin{cases}
   1,& \text{if } y_{ip}=y_{iq}\\
   0,& \text{otherwise}
\end{cases}
\label{B_definition}
\end{equation}
\noindent where $l_{pc}$ is the line segment connecting pixel $p$ to the object center $c$ and $q$ is any pixel incident on line $l_{pc}$. $L_{sh}$ is trained to assign to all such $q$ pixels a label identical to the label of pixel $p$ as long as (i) $p$ and $q$ have the same ground truth labels ($B_{pq}^{i}=1$) and (ii) the difference between the ground truth label and the predicted labels for $p$ is non-zero ($|y_{ip}-P(y_{ip}|X(i);\theta)|>0$). \zm{The 3rd term of~\eq{\ref{star_shape_loss}} determines how labels of pixels internal to the lesion are penalized to ensure star shapes, whereas the first two terms of~\eq{\ref{star_shape_loss}} are designed to allow discontinuities of pixel labels across the ground truth boundary of the lesion and ignore the star shape term when the given label is true.} In \fig{\ref{starshape}}-(c), $p=p_3$ and $p=p_4$ are examples where the value of $\sum_{q \in l_{pc}}|P(y_{ip}|X(i);\theta)-P(y_{iq}|X(i);\theta)|$ is positive while their assigned labels should not be penalized.  Condition (i) chooses a set of pixels $q$ on $l_{p_3c}$ and allows discontinuities between the background ($p_3$) and foreground assigned labels and, condition (ii) enforces the loss function not to penalize the label assigned to $p_4$.

In our implementation of \ref{star_shape_loss}, instead of penalizing the difference between the predicted probabilities and ground truth labels for all the points on the straight line $l_{pc}$, we only examine the $m$ closest pixels to $p$ on $l_{pc}$ and compute the loss value per pixel $p$ based on those $m$ predicted probabilities. We also quantize, to a set of $d$ directions, the possible angles of all lines passing through $p$.  \cmmnt{encoded into $8$ different kernels of size $k\times k$, where $k=2m+1$.} 
\cmmnt{Finite violation cost of the star shape loss term $L_{sh}$ is used to make the star shape a soft (nor hard) constraint that we would like to meet but not at the cost of violating the cross-entropy loss.}\cmmnt{Unlike the implementation of star shape prior in classical energy-based minimization approaches~\cite{veksler2008star,nosrati2015segmentation},} In the training of our deep network, we automatically find the star object center from binary ground truth maps. At inference time, we do not need to supply the center of star objects as prediction maps are achieved by a forward pass through the network whose parameters are already trained to generate segmentations.\\

\section{Experiments}

\begin{figure*}[t]
\centering
\includegraphics[width=4.2in]{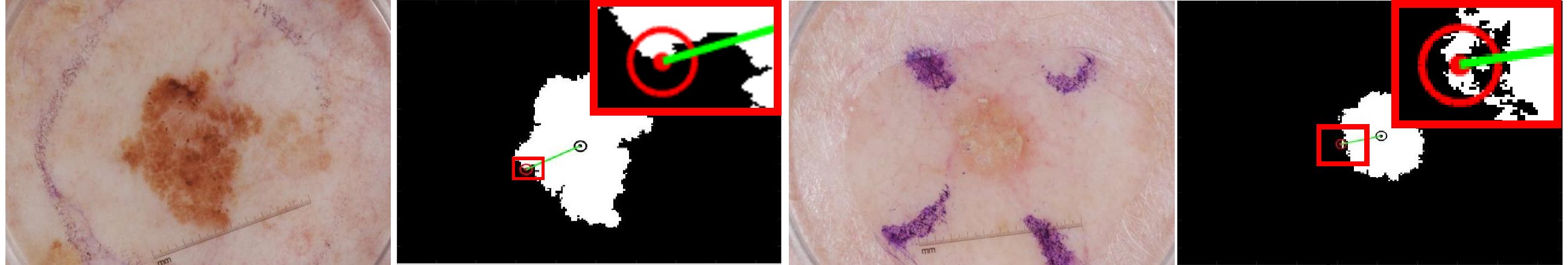}
\caption{Examples of skin lesion pixels violating the star shape constraint.}
\vspace{-1.5em}
\label{nonstar_samples}
\end{figure*}

\noindent \textbf{Data description}~~We validated our proposed segmentation approach on dermoscopy data provided by the International Skin Imaging Collaboration (ISIC) at ISBI $2017$ \emph{Skin Lesion Analysis Towards Melanoma Detection Challenge}~\cite{codella2017skin}. The data set contains $2000$ training, $150$ validation, and $600$ test images.\cmmnt{$8$-bit RGB images paired with the same size groundtruth binary masks... where pixel values of $255$ indicate skin lesions and values of 0 represents normal skin. Each ground truth binary mask was prepared by an expert that manually drew the lesion boundaries. We used the same split of training and validation data as provided by challenge organizers to set network hyper-parameters.} We first re-scaled all images to $192\times192$ pixels and normalized each RGB channel by the mean and standard deviation of the training data. To confirm the suitability of adopting the star-shape prior for this task, we calculated the percentage of segmentation mask pixels that violate the star shape definition to be only 0.14\% over the whole dataset (0.05\% of training, 0.3\% of validation, and 0.38\% of test image pixels).
\fig{\ref{nonstar_samples}} shows examples of rare pixels where the star shape constraint is violated.\\

\noindent \textbf{Network architecture}~~We exploited two state-of-the-art fully convolutional network architectures to evaluate our proposed new loss: 1)U-Net\cite{ronneberger2015u} 2)ResNet-DUC. ResNet-DUC deploys the FCN version of ResNet-152, pretrained on ImageNet as an encoder~\cite{he2016deep}. Instead of using multiple deconvolutional layers to decode low resolution feature maps into the original image size prediction maps, single Dense Upsampling Convolution (DUC) layer is used to reconstruct fine-detailed information from coarse feature maps~\cite{wang2017understanding}. Furthermore, dilated convolutions are used in the encoder to benefit from multi-scale contextual information from previous layers activations~\cite{yu2015multi}.\\

\noindent \textbf{Implementation}~~\cmmnt{We implemented all deep models with the PyTorch library.} We trained deep networks implemented with the PyTorch library, over mini-batches of size 12.\cmmnt{NVIDIA Titan X  Using the validation data and their corresponding ground truth, in addition to By a grid search, w} We tuned all hyper-parameters on the validation set. Loss functions are optimized using the stochastic gradient descent algorithm with an initial learning rate of $10^{-4}$. The learning rate was divided by $10$ when the performance of model on validation data set stopped improving. Momentum and weight decay were set to $0.99$ and $5\times10^-5$, respectively.  For the implementation of the star shape regularized loss function, $\alpha=1$, $\beta=5$, $m=6$ and $d=8$. We first trained the deep network with binary cross entropy function for $5$ epochs and then fine-tuned the network parameters with the proposed loss function. \zm{Training takes ~2 days and test takes ~1 sec/image on our 12 GB GPU.}\\

\noindent \textbf{Results}~~We evaluated the performance of U-Net and ResNet-DUC trained with and without the star shape prior. As shown in \tab{\ref{qnt_results}}, using our shape regularized loss function in the training of U-Net and ResNet-DUC, the Jaccard index is improved by more than $3\%$ (row A vs. B and row C vs. D). We measured the statistical significance of our results by exploring the Jaccard index over the test data. We used the non-parametric Wilcoxon signed rank sum test and found that the results of U-Net and ResNet-DUC with and without incorporation of star shape prior are statistically significantly different at $p < 0.05$.
\cmmnt{We evaluated the performance of U-Net and ResNet-DUC trained with and without the star shape prior. As shown in \tab{\ref{qnt_results}}, using our shape regularized loss function in the training of U-Net and ResNet-DUC, we obtained a statistically significant improvement in the Jaccard index of more than $3\%$
($p<0.05$ using the non-parametric Wilcoxon signed rank sum test).}

\begin{figure*}[!ht]
\centering
\vspace{-1em}
\includegraphics[width=3.8in]{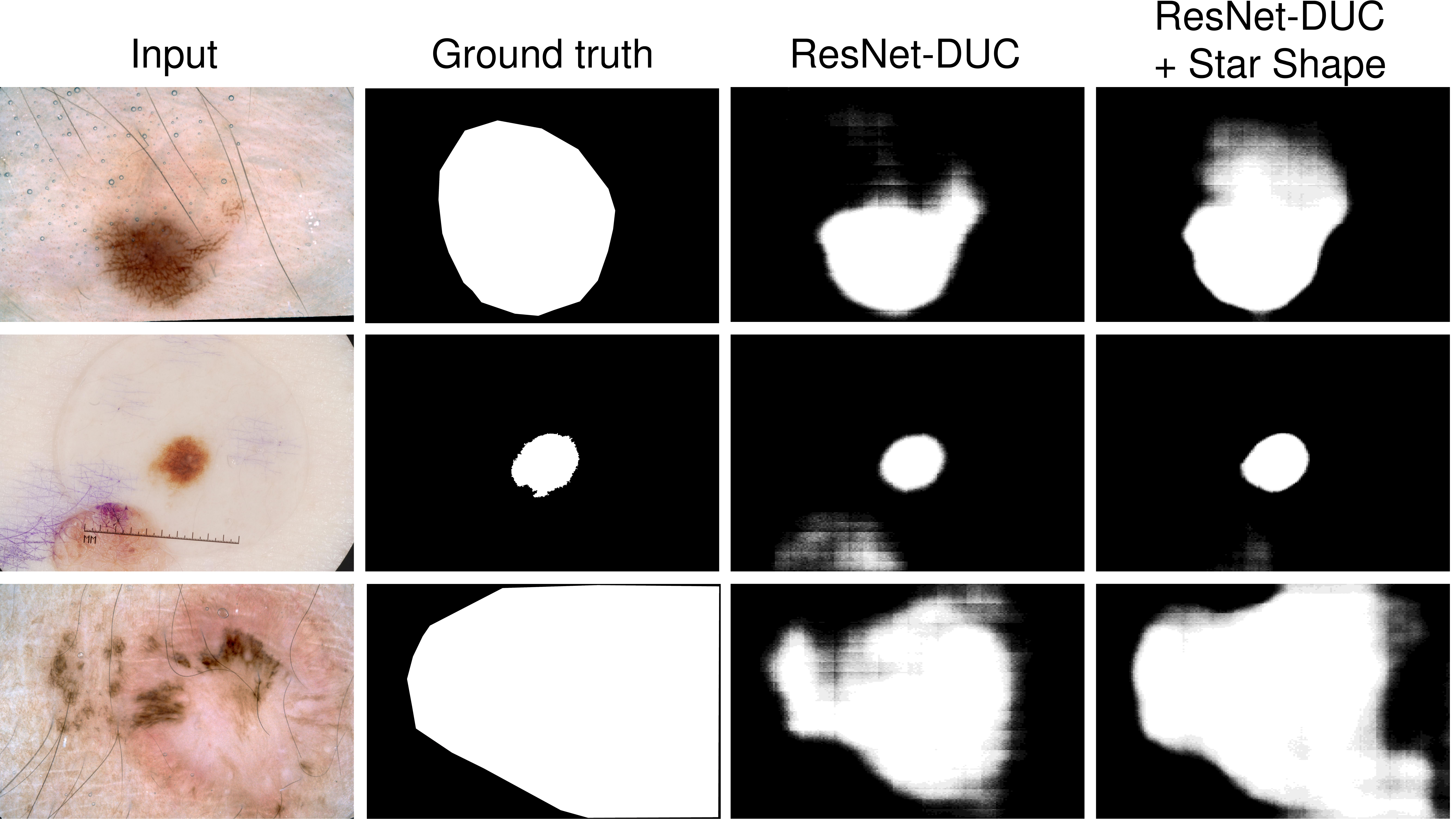}
\caption{Qualitative comparison of ResNet-DUC architecture results with and without star shape prior.}
\vspace{-1em}
\label{qlt_results}
\end{figure*}

We compared our proposed method with $21$ competing methods participating in the challenge. The ResNet-DUC architecture trained with our star shape regularized loss achieved the first rank based on the challenge ranking metric, Jaccard index. \tab{\ref{qnt_results}}, rows E, F and G, show results of the first three ranked teams. Although all top three teams used FCNs to perform image segmentation, in contrast to our work, they employed various additional steps like averaging over multiple model results, multi-scale image input as well as pre- and post-processing approaches like inclusion of different color spaces in the input and multi-thresholding.
Qualitative results of our proposed approach are presented in \fig{\ref{qlt_results}}. Encoding star shape prior into the loss function results in smoother prediction maps with a single connected component as lesion for most cases.

\begin{table*}[t!]
\centering

\caption{Segmentation quantitative performance. Bold numbers indicate the best performance. All values are in percentages.}
\vspace{-1em}
\begin{center}
\resizebox{1\textwidth}{!}{
\begin{tabular}{|c||l||c|c|c|c|c|}
\hline
\hline
\textbf  &Method & Jaccard&Dice~~& Accuracy & Specificity & Sensitivity\\
\hline
\hline
A & U-Net~\cite{ronneberger2015u}& 70.5 & 79.7 & 91.8 & 97.8 & 77.0\\
\hline
B & U-Net + Star Shape & 73.3 & 82.4 & 92.4 & 95.3 & 85.4 \\
\hline
C & ResNet-DUC~\cite{wang2017understanding}& 74.0 & 83.3 & 93.00 & 98.2 & 80.0 \\
\hline
D &  ResNet-DUC + Star Shape& \textbf{77.3} & \textbf{85.7} & \textbf{93.8} & 97.3 & \textbf{85.5} \\
\hline
E & Yuan et al.~\cite{yuan2017automatic_1}& 76.5 & 84.9 & 93.4 & 97.5  & 82.5\\
\hline
F & Berseth et al.~\cite{berseth2017isic_2} & 76.2 & 84.7 & 93.2 & 97.8 & 82.0 \\
\hline
G & Bi et al.~\cite{bi2017automatic_3} & 76.0& 84.4 & 93.4 & \textbf{98.5} & 80.2 \\
\hline
\hline
\end{tabular}}
\end{center}
\label{qnt_results}
\end{table*}

\cmmnt{
\begin{table*}[ht!]
\centering
\caption{ISBI 2017 segmentation challenge results. All values are in percentages.}
\vspace{-1em}
\begin{center}
\resizebox{1\textwidth}{!}{
\begin{tabular}{|c||l||c|c|c|c|c|}
\hline
\hline
\textbf  &Method & Jaccard&Dice~~& Accuracy & Specificity & Sensitivity\\
\hline
\hline
A & Yuan et al.~\cite{yuan2017automatic_1}& \textbf{76.5} & \textbf{84.9} & \textbf{93.4} & 97.5  & \textbf{82.5}\\
\hline
B & Berseth et al.~\cite{berseth2017isic_2} & 76.2 & 84.7 & 93.2 & 97.8 & 82.0 \\
\hline
C & Bi et al.~\cite{bi2017automatic_3} & 76.0& 84.4 & \textbf{93.4} & \textbf{98.5} & 80.2 \\
\hline
\hline
\end{tabular}}
\end{center}
\label{challenge_results}
\vspace{-2em}
\end{table*}
}

\section{Conclusion}

We encoded the star shape prior in the loss function of an end-to-end trainable fully convolutional network to generate more accurate and plausible skin lesion segmentations. In contrast to energy minimization approaches, our proposed framework does not require computationally expensive optimization at inference time nor a user-defined object centre. Our experiments indicated that leveraging the prior knowledge in fully convolutional networks yield convergence to an improved output space. In future works, we will extend to other prior information including but not limited to anatomically meaningful priors in fully convolutional networks trained for other 2D and 3D medical imaging applications.\\

\noindent\textbf{Acknowledgments.} We gratefully thank NVIDIA Corporation for the donation of the Titan X GPU used for this research.

\bibliographystyle{splncs03}
\bibliography{ref}

\end{document}